\documentclass{article}

\usepackage[utf8]{inputenc}

\usepackage{microtype}
\usepackage{graphicx}
\usepackage{subfigure}
\usepackage{booktabs} % for professional tables

\usepackage{amsmath,graphicx}
\usepackage{enumitem}
\usepackage{xcolor}
\usepackage{mathtools}
\usepackage{cite}
\usepackage{setspace}
\usepackage{wrapfig}
\usepackage[hyphens]{url}
\usepackage[hidelinks]{hyperref}
\usepackage{cleveref}
\usepackage{enumitem}

\DeclareMathOperator*{\argmax}{arg\,max}

\usepackage[accepted]{icml2020}

\icmltitlerunning{A study of latent monotonic attention variants}

\begin{document}

\twocolumn[
\icmltitle{A study of latent monotonic attention variants}



\icmlsetsymbol{equal}{*}

\begin{icmlauthorlist}
\icmlauthor{Albert Zeyer}{rwth,apptek}
\icmlauthor{Ralf Schlüter}{rwth,apptek}
\icmlauthor{Hermann Ney}{rwth,apptek}
\end{icmlauthorlist}
  
\icmlaffiliation{rwth}{Human Language Technology and Pattern Recognition, Computer Science Department, RWTH Aachen University, Aachen, Germany}
\icmlaffiliation{apptek}{AppTek GmbH, Aachen, Germany}
  
%\icmlcorrespondingauthor{Albert Zeyer}{zeyer@cs.rwth-aachen.de}
\icmlcorrespondingauthor{}{\texttt{\{zeyer, schlueter, ney\}@cs.rwth-aachen.de}}
    
\icmlkeywords{Soft attention, hard attention, monotonicity, segmental models, latent models}
  
\vskip 0.3in
]  



\printAffiliationsAndNotice{}  % leave blank if no need to mention equal contribution

% \begin{abstract}
% This document provides a basic paper template and submission guidelines.
% Abstracts must be a single paragraph, ideally between 4--6 sentences long.
% Gross violations will trigger corrections at the camera-ready phase.
% \end{abstract}

% http://tex.stackexchange.com/questions/10684/vertical-space-in-lists
% package enumitem
\setlist{
	itemsep=4pt plus 2pt minus 2pt,
	parsep=2pt plus 2pt minus 1pt,
	topsep=2pt plus 2pt minus 2pt,
	partopsep=2pt plus 2pt minus 2pt}

\begin{abstract}
End-to-end models reach state-of-the-art performance
for speech recognition,
but global soft attention is not monotonic,
which might lead to convergence problems, to instability,
to bad generalisation,
cannot be used for online streaming,
and is also inefficient in calculation.
Monotonicity can potentially fix all of this.
There are several ad-hoc solutions or heuristics to introduce monotonicity,
but a principled introduction is rarely found in literature so far.
In this paper, we present a mathematically clean solution to introduce monotonicity,
by introducing a new latent variable which represents the audio position or segment boundaries.
We compare several monotonic latent models to our global soft attention baseline
such as a hard attention model, a local windowed soft attention model,
and a segmental soft attention model.
We can show that our monotonic models perform as good as the global soft attention model.
We perform our experiments on Switchboard 300h.
We carefully outline the details of our training and release our code and configs.
\end{abstract}
%

%\vspace{\baselineskip}
\section{Introduction}
\label{sec:intro}

The conventional hidden Markov model (HMM),
including the hybrid neural network (NN) / HMM
\cite{bourlard1990continuous,robinson1994hybridrnn}
is a \emph{time-synchronous} model, which defines a probability per input frame
(either $p(x_t|y_t)$ or $p(y_t|x_t)$).
As such, it implicitly enforces monotonicity.
Connectionist temporal classification (CTC) \cite{graves2006ctc}
can be seen as a special case of this \cite{zeyer2017:ctc}.
Generalisations of CTC are the recurrent neural network transducer (RNN-T) \cite{graves2012rnnt,battenberg2017rnnt}
or the recurrent neural aligner \cite{sak2017rna,dong2018rna}
or further extended generalized variations of the transducer \cite{zeyer2020:transducer}.

The encoder-decoder-attention model \cite{bahdanau2014globalatt}
is becoming one of the most competitive models
for speech recognition
\cite{chiu2018sotaatt,zeyer2018:asr-attention,park2019specaugment,tuske2020attswb}.
This model directly models the posterior probability $p(y_1^N | x_1^T)$
for the output labels $y_1^N$ (word sequence),
given the input sequence $x_1^T$ (audio features).
This is factorised into $\prod_i p(y_i | y_1^{i-1}, x_1^T)$.
We call this a \emph{label-synchronous model} (sometimes also called \emph{direct model})
because each NN output corresponds to one output label.
This is in contrast to the time-synchronous models like hybrid NN/HMM, CTC and RNN-T.
The encoder-decoder (label-sync.) model usually uses global soft attention in the decoder.
I.e.~for each output label $i$, we access the whole input sequence $x_1^T$.
This global soft attention mechanism is inefficient (time complexity $O(T \cdot N)$)
and can violate monotonicity.

Monotonicity is necessary to allow for online decoding.
Also, for monotonic sequence-to-sequence tasks like speech recognition,
a model which is enforced to be monotonic might converge faster,
and should be more stable in decoding.
In this work, we want to introduce monotonicity in a mathematically clean way
by introducing a \emph{latent variable} which presents the position in the audio (or frame-wise encoder).

\section{Related work}

Several approaches exists to introduce monotonicity for label-synchronous models.
We categorize these different approaches and outline their advantages and disadvantages.

\subsection{Soft constraints: By training, or by attention energy tendency}
Monotonicity can be encouraged by the model
(e.g.~by modeling the attention energy computation in such a way that it tends to be monotonic)
or during training (by additional losses, or some initialization with linear alignment)
\cite{tachibana2018guidedatt,zhang2018forwardatt}.
While these help for convergence,
these are soft constraints, or just guidance,
i.e.~the model still can violate monotonicity.

\subsection{Hybrids of attention and CTC/RNN-T}

The idea is that a time-synchronous model like
CTC or RNN-T enforces the monotonicity,
and the attention model contributes for better performance
\cite{watanabe2017hybridctcatt,hori2017jointctcatt,kim2017jointctcatt,
moritz2019triggeredatt,moritz2019jointctcatt,
miao2019mocha,chiu2019twopass}.
Often there is a shared encoder, and one CTC output layer
and another common decoder with attention,
and the decoding combines both the CTC and the attention decoder.
The decoding can be either implemented in a time-synchronous way
or in a label-synchronous way.
Depending on the specific details of the model and decoding,
monotonicity is strictly enforced or just strongly encouraged
(but it could happen that the attention scores dominate over the CTC model or so).
These models usually perform quite well.
However both the model and the decoding procedure become more complicated,
and often rely on several heuristics.
And it can be seen as a model combination,
which makes the performance comparison somewhat unfair to single models.

\subsection{Deterministic methods}
\label{sec:deterministic-monotonic}
Monotonicity can be introduced in a more conceptional way
via deterministic methods,
which model explicitly the source position
\cite{graves2013gaussatt,luong2015globalatt,%
tjandra2017local,hou2017gaussian,%
jaitly2016online,raffel2017online,%
chiu2017mocha,fan2018mocha,miao2019mocha,%
merboldt2019:local-monotonic-att,he2019stepwise,
zhang2019windowedatt}.
These approaches all provide strict monotonicity by using some \emph{position scalar}
(e.g.~for a running window over the encoder output),
and it is enforced that the position can only increase.
In these approaches, the position is not interpreted as a \emph{latent variable},
but it is rather calculated by a \emph{deterministic function}
given the current state of the decoder.
These can be fully differentiable (if the position is a real number scalar)
or rely on heuristics/approximations (if the position is a discrete number).
This includes the implicit assumption that there exists a deterministic method
which determines the position for the next label.
We argue that this is a too strong assumption.
In practice, given some history of input (audio) and previous labels (words),
the model cannot tell for sure where the next label will occur.
The model eventually can only estimate a probability or score.
And later in decoding, after a few more input frames or words,
we might see that the model did a wrong choice for the position.
The beam search should be able to correct this,
which is not possible if it is simply a deterministic function.
If it cannot correct it, then the performance will have a bottleneck at this deterministic function,
because when its estimate is bad or wrong, the further decoding will be affected consequently.
\emph{MoChA} \cite{chiu2017mocha} is probably one of the most prominent examples in this category.
As it still does soft attention within one chunk, it can compensate bad chunk positions to some degree
but not fully.
% chiu2017mocha: match the performance of a model using an offline soft attention mechanism
% fan2018mocha: LC-AMoChA, which has only 3.5% relative performance reduction to LAS baseline on our internal Mandarin corpus.
%The mathematically clean way to allow that is to introduce the position as a latent variable.
%This is actually a generalization of these deterministic approaches, as we will see later. 

\subsection{Latent variable models}

A variety of works have introduced an explicit \emph{latent variable} to model
the alignment.
We always have and even need this for hybrid NN/HMM models, CTC and RNN-T.
%For direct / label-synchronous models, it is not necessarily needed,
%thus not all models have it, like the standard encoder-decoder-attention model.
One example of early label-synchronous models which have such a latent variable
are the segmental conditional random fields (CRF)
\cite{ostendorf1996segmental,zweig2009segmental,lu2016segmental}.
The latent variable is usually discrete, and often represents the position in the encoder,
or the segment boundaries.
We will categorize the model for the latent variable by dependency orders in the following.

\subsubsection{0-order models}

The probability distribution of the latent variable can be independent from its history (0 order)
\cite{bahuleyan2017variational,%
shankar2018hardatt,shankar2018posterioratt,deng2018latent,wu2018hard}.
The marginalisation over the latent variable becomes simple and efficient
in that case
and can easily be done in both training and decoding.
However, it does not allow for strong alignment models,
and also does not allow to constrain for monotonicity,
because that would add a dependence on the history and then it is not 0-order anymore.

\subsubsection{1st order models}
If the probability distribution of the latent variable depends only on its predecessor,
%but not further in its history,
then it is of 1st order
\cite{yu2016online,yu2016noisychannel,%
wang17:directhmm,wwang18:neuralhmm,%
alkhouli16:align,%
beck2018:segmental,beck2018:alignment}.
A first-order dependency allows to directly enforce monotonicity.
The full marginalisation can still be calculated efficiently for 1st order models
using dynamic programming \cite{yu2016online,wwang18:neuralhmm}

\subsubsection{Higher order models}
The probability distribution of the latent variable
can also depend on more than just the predecessor,
or even on the full history
\cite{mnih2014visatt,ba2014visatt,xu2015show,%
vinyals2015pointernets,%
lawson2018hard,arivazhagan2019milk,%
alkhouli17:align,alkhouli18:trafoalign}.
This is often the case
if some recurrent decoder is used and the probability distribution depends on that state,
which itself depends on the full history.
The full marginalisation becomes infeasible for higher order,
thus requires further approximations such as the maximum approximation
\cite{alkhouli17:align,alkhouli18:trafoalign}.

\subsubsection{Decoding with first or higher-order latent variable models}

Only few other work includes the latent variable in beam search during decoding
\cite{yu2016online,yu2016noisychannel,
wwang18:neuralhmm,
alkhouli16:align,alkhouli17:align,alkhouli18:trafoalign,
beck2018:segmental}
while most other work uses sampling or a greedy deterministic decision during decoding,
which again leads to all the problems outlined in \Cref{sec:deterministic-monotonic}.
We argue that we should properly search over the latent variable space
and incorporate that into the beam search decoding procedure
such that the decoding can correct wrong choices on the latent variable.
Also this is mostly straight forward and mathematically clean.

Note that if we simplify the beam search to simply pick the $\argmax$ for the latent variable
in each decoder step, we get back to deterministic models.
In that sense, latent variable models are a strict generalization of all the deterministic approaches.

\section{This work: our latent variable model}

We conceptually introduce monotonicity in a mathematically clean way
to our label-synchronous encoder-decoder model
by a discrete \emph{latent variable}
which \emph{represents the position} (or segment boundary) in the audio (or frame-wise encoder).
We now have several options for the model assumptions.
For the monotonicity, we at least need a model of 1st order.
We make it \emph{higher order} and depend on the full history.
This is is a trade-off between a more powerful model
and more approximations needed in decoding and training,
such as the maximum approximation.
The maximum approximation has successfully and widely been used for speech recognition
in different scenarios such as decoding and Viterbi training for hybrid NN/HMM models
\cite{yu2014asrbook}.
This is why we think the maximum approx.~should be fine for label-sync.~models as well.
And on the other side, a more powerful model could improve the performance.

%It is similar to other 1st/higher order ? ... \TODO

We argue that these models with a latent variable are conceptually better founded
than the deterministic models which rely on heuristics
without good mathematical justification.
In addition, in such an approach,
the conventional maximum approximation can be naturally introduced
to achieve efficient and simple training and decoding.

As we want to understand the effect of this latent variable, and the hard monotonicity constrain on it,
we want to keep the differences to our
baseline encoder-decoder model with global soft attention
as small as possible.
This means that we still have a
bidirectional long short-term memory (LSTM) \cite{hochreiter1997lstm} encoder
and an offline feature extraction pipeline is offline.
%, i.e.~the model is not fully online capable yet,
There is a wide range of existing work on making these missing parts online capable
\cite{zeyer16:onlinebidir,peddinti2017online,xue2017lcblstm}.
%which is out of the scope for this work.
%We skip this in this work to focus on the properties of introducing the latent variable.
%We study multiple variants where we have introduced a hard constraint on monotonicity,
%at least conceptually.
In all our latent models, the added latent variable represents either:
\begin{itemize} %[itemsep=0pt]
\item a single discrete position in the encoder output,
\item a discrete center position of a window over the encoder output,
\item or a segment boundary on the encoder output.
\end{itemize}
To further keep the difference to the baseline small,
the probability of the latent variable is simply given by the attention weights
(but masked to fulfil monotonicity).
Also the remaining part of the decoder is kept exactly the same,
except of how we calculate the attention context,
which is not by global soft attention anymore.
We do not improve on the runtime complexity in this work,
to stay closer to the baseline for a better comparision.
However, it is an easy further step to make the runtime linear, as we will outline later.

Also, in the existing works,
the final model properties are rarely discussed beyond the model performance.
We perform a \emph{systematic study} of different properties of our latent variable models,
considering modeling, training, and decoding aspects.

The experiments were performed with RETURNN \cite{zeyer2018:returnn},
based on TensorFlow \cite{tensorflow2015}.
%\ifdefined\isaccepted
All the code and the configs for all our experiments is publicly available%
\footnote{\scriptsize\url{https://github.com/rwth-i6/returnn-experiments/tree/master/2021-latent-monotonic-attention}}.
%\else
%We publish all the code and configs of our experiments as part of the supplementary material,
%and provide a link in the accepted version.
%\fi

\section{Models}

We
\ifdefined\isaccepted
closely follow the
\else
use a
\fi
LSTM-based global soft attention model
\ifdefined\isaccepted
as defined in
\cite{zeyer2018:asr-attention,zeyer2018:attanalysis,zeyer2019:trafo-vs-lstm-asr}.
\else
similar to \cite{bahdanau2014globalatt,luong2015globalatt,
chorowski2015attention,chan2016las,bahdanau2016endtoendasr}.
\fi
In all cases, the model consists of an encoder,
which learns a high-level representation of the input signal $x_1^T$,
\[ h_1^{T'} =  \operatorname{Encoder}(x_1^T) . \]
We apply time downsampling, thus we have $T' \le T$.
In our case, it is a bidirectional LSTM with time-max-pooling.
The global soft attention model defines the probability
\[ p(y_1^N | x_1^T) = \prod_{i=1}^N p(y_i | y_1^{i-1}, x_1^T) \]
with
\[ p(y_i | y_1^{i-1}, x_1^T) = p(y_i | y_1^{i-1}, h_1^{T'}) = p(y_i | y_1^{i-1}, c_1^i), \]
for $i \in \{1, \dots, N\}$,
where $c_i$ is the attention context vector.
We get the dependence on the full sequence $c_1^i$ because we use a recurrent NN.

\subsection{Baseline: Global soft attention}

There is no latent variable.
The attention context $c_i$ is calculated as
\[ c_i = \sum_{t=1}^{T'} \alpha_{i,t} (y_1^{i-1}, \alpha_1^{i-1}, c_1^{i-1}) \cdot h_t, \]
and the attention weights $\alpha_i$ are a probability distribution over $\{1, \dots, T'\}$,
which we calculate via MLP-attention \cite{bahdanau2014globalatt,luong2015globalatt}.
In this model, there is no explicit alignment between an output label $y_i$ and the input $x_1^T$
or the encoder output $h_1^{T'}$.
However, the attention weights can be interpreted as a soft alignment.
The model also uses weight feedback \cite{tu2016weightfeedback}.

The model is trained by minimising the negative log probability over the training dataset $\mathcal{D}$
\[ L = - \sum_{(x,y) \in \mathcal{D}} \sum_{i=1}^N  \log p(y_i |  y_1^{i-1}, x_1^T) . \]
We also use a pretraining scheme such as growing the model size,
which we outline in the experiments section.
Decoding is done by searching for
\[ \argmax_{N,y_1^N} \sum_{i=1}^N  \log p(y_i |  y_1^{i-1}, x_1^T) \]
which is approximated using label-synchronous beam search.

\subsection{Latent models}

When introducing a label-synchronous discrete latent variable $t_i$ for all $i \in \{1,\dots,N\}$,
we get
\begin{align*}
p(y_1^N | x_1^T) &= \sum_{t_1^N}  p(y_1^N, t_1^N | x_1^T)  \\
&= \sum_{t_1^N} \prod_{i=1}^N p(y_i, t_i | y_1^{i-1}, t_1^{i-1}, x_1^T) \\
&= \sum_{t_1^N} \prod_{i=1}^N p(y_i | y_1^{i-1}, t_1^i, x_1^T) p(t_i | y_1^{i-1}, t_1^{i-1}, x_1^T) . %\\
%&\approx \max_{t_1^N} \prod_{i=1}^N p(y_i | y_1^{i-1}, t_1^i, x_1^T) p(t_i | y_1^{i-1}, t_1^{i-1}, x_1^T)
\end{align*}
Analogous to the baseline, we use
\[ p(y_i | y_1^{i-1}, t_1^i, x_1^T) = p(y_i | y_1^{i-1}, c_1^i), \]
where the attention context $c_i$ depends also on $t_i$ now.

%For all the strict monotonic models,
%we introduce a new latent variable $t_i \in \{1,\dots,T'\}$ for all $i$,
We construct the model in a way that the access to $h_1^{T'}$ via $t_i$ is monotonic
by $t_i \le t_{i+1}$ or even $t_i < t_{i+1}$.
In every case, the model is trained again by minimising the negative log probability,
and using the max.~approximation for $\sum_{t_1^N}$,
\[ L = - \sum_{(x,y) \in \mathcal{D}} \max_{t_1^N} \sum_{i=1}^N  \log p(y_i, t_i |  y_1^{i-1}, t_1^{i-1}, x_1^T) . \]
The $\argmax_{t_1^N}$ is further approximated by beam search,
which is also called \emph{forced alignment} or \emph{Viterbi alignment}.
Also, in all cases, we try to be close to the global soft attention model,
such that a comparison is fair and direct,
and even importing model parameters is reasonable.

The probability for the latent variable $t_i$ is simply given by the attention weights
\[ p(t_i = t| y_1^{i-1}, t_1^{i-1}, x_1^T ) = \alpha'_{i,t} \]
with $\alpha'_{i,t} = 0 \quad \forall t < t_{i-1}$.
The attention weights in this case are masked to fulfil monotonicity
($t < t_{i-1}$ if we want strict monotonicity, else $t \le t_{i-1}$),
and $\alpha'_{i,t} \propto \alpha_{i,t}$ renormalised accordingly.
We choose this model $p(t|\dots)$ based on the attention weights
to keep a strong similarity to the baseline model.
However, because of this model, we still have $O(T\cdot N)$ runtime complexity.
In future work, we can further deviate from this,
and compare to other models for $p(t|\dots)$
which allow for linear runtime $O(T+N)$ or even $O(T)$.
E.g.~an obvious alternative is to use a Bernoulli distribution
and to define it frame-wise, not globally over all frames.

In all latent models, we define the attention context $c_i$ as
\begin{align}
c_i = \sum_{t=w_i}^{w'_i} \alpha''_{i,t} \cdot h_t, \label{eq:latent-window}
\end{align}
i.e.~we use soft attention on a window $[w_i,w'_i]$
(or hard, if it is a single frame), for some $\alpha''$ as described in the following.

Decoding is done by searching for
\[ \argmax_{N,y_1^N,t_1^N} \sum_{i=1}^N  \log p(y_i, t_i |  y_1^{i-1}, t_1^{i-1}, x_1^T) \]
which is approximated using label-synchronous beam search.

\subsection{Monotonic hard attention}

By simply setting $[w_i,w'_i] = [t_i,t_i]$ in \Cref{eq:latent-window},
we have the attention only on a single frame,
and simply $\alpha''_{i,t} = 1$.
This can be interpret as hard attention instead of soft attention,
and $t_i$ is the position in $h_1^{T'}$.
Everything else stays as before.
There are multiple motivations why to use hard attention instead of soft attention:
\begin{itemize}
\item The model becomes very simple.
\item In the global soft attention case,
we can experimentally see that the attention weights are usually very sharp,
often focused almost exclusively on a single frame.
\end{itemize}

\subsection{Monotonic local windowed soft attention}

We use $[w_i,w'_i] = [t_i - D_l, t_i + D_r]$
for fixed $D_l, D_r$.
In this case, the latent variable $t_i$ is the center window position,
and $\alpha''_{i,t} \propto \alpha_{i,t}$ renormalised on the window.
This model is very similar to \cite{merboldt2019:local-monotonic-att}
but with a latent variable.

\subsection{Monotonic segmental soft attention}

Here we make use of the latent variable $t_i$ as the segment boundary,
i.e.~we use $[w_i,w'_i] = [t_{i-1}+1, t_i]$.
Here, we have a separate model for $\alpha''_{i,t}$,
which are separately computed MLP soft attention weights.

\section{Training procedure for latent models}
\label{sec:training-latent}

For one training sample $(x,y) \in \mathcal{D}$,
the loss we want to minimise is
\[ L = - \max_{t_1^N} \log p(y_i^N, t_i^N | x_1^T) , \]
i.e.~we have to calculate $\arg\max_{t_1^N}$,
i.e.~search for the best $t_1^N$,
which we call \emph{alignment}.
This is the \emph{maximum approximation} for training.
This can be done using dynamic programming (beam search, Viterbi).
%Because of the high-order dependency in $t$,
%we cannot do recombination on partial paths in that search,
%and thus must use beam search.
We could also decouple the search for the best alignment
from updating the model parameters,
but in our experiments, we always do the search online on-the-fly,
i.e.~in every single mini-batch, it uses the current model for the search.
We note that this beam search
is exactly the same beam search implementation which we use during decoding,
but we keep the ground truth $y_1^N$ fixed,
and only search over the $t_1^N$.

For stable training, we use the following tricks,
which are common for other models as well, but adopted now for our latent variable models.
% hard-att3.textend.trecombtrain.attwl01.base2.conv2l.specaug4a.ls01.laplace1000
% hard-att-imp3.textend.trecombtrain.attwl01.base2.conv2l.specaug4a.ls01.laplace1000
\begin{itemize}
\item The maximum approximation can be problematic in the very beginning of the training,
when the model is randomly initialized.
In the case of HMM models, it is common to use a linear alignment in the beginning,
instead of calculating the Viterbi alignment ($\argmax_{t_1^N}$).
We do the same for our latent variable models.
\item We store the last best alignment (initially the linear alignment) and its score.
When we calculate a new Viterbi alignment, we compare the score to the previous best score,
and update the alignment if the score is better.
This greatly enhances the stability early in training,
esp.~when we still grow the model size in pretraining.
\item We can use normal global soft attention initially.
Note that this is still slightly different compared to the global soft attention baseline
because of the other small differences in the model, such as different (hard) weight feedback.
\end{itemize}

%However, for stability reasons, for the first training epoch,
%we use a linear alignment instead.
%In addition to that, we always store the most recent best alignment,
%according to the score $p(y_i^N, t_i^N | x_1^T)$.
%We only use the new alignment if its score is better,
%and otherwise fall back to the best most recent alignment.

Given an existing alignment $t_1^N$ for this training example,
the loss becomes
\[ L = \sum_{i=1}^N -\log p(y_i |  y_1^{i-1}, c_1^i) - \log p(t_i | y_1^{i-1}, c_1^{i-1}) . \]
We note that this is the same loss as before,
with the additional negative log likelihood for $p(t_i|\dots)$.
We can also easly introduce a different loss scale for each log likelihood.
We use the scale $0.1$ by default for $p(t_i|\dots)$.

% baseline: average epoch time 'GeForce GTX 1080 Ti': 0:30:20
% hard att: average epoch time 'GeForce GTX 1080 Ti': 0:42:10 | 0:41:53 (+40%)
% hard segm: average epoch time 'GeForce GTX 1080 Ti': 1:10:53 (>+100%)
% hard local win: average epoch time 'GeForce GTX 1080 Ti': 1:04:56 (>+100%)
As we do the search for the optimal $t_1^N$ on-the-fly during training,
this adds to the runtime and makes it slower.
However, on the other side, the model itself can be faster, depending on the specific implementation.
In our case, for the hard attention model,
we see about 40\% longer training time.
If we would use a fixed alignment $t_1^N$, it would actually be faster to train than the baseline.

\section{Decoding with latent variable}
\label{sec:decoding}

The decoding without an additional latent variable is performed as usual,
using label synchronous beam search with a fixed beam size
\cite{sutskever2014seq2seq,bahdanau2014globalatt,zeyer2018:returnn}.
When we introduce the new latent variable into the decoding search procedure,
the beam search procedure stays very similar.
We need to search for
\[ \argmax_{N,y_1^N,t_1^N}
\sum_{i=1}^N \log p(y_i |  y_1^{i-1}, c_1^i) + \log p(t_i | y_1^{i-1}, c_1^{i-1}) . \]
As $p(y_i|\dots)$ depends on $t_i$ (via $c_i$),
in any decoder step $i$,
we first need to hypothesize possible values for $t_i$,
and then possibe values for $y_i$,
and then repeat for the next $i$.
We have two beam sizes, $K_y$ and $K_t$
for $p(y|\dots)$ and $p(t|\dots)$ respectively.

%How do we expand on possible choices for the latent variable $t_i$.
The expansion of the possible choices for $t_i$ is simply
analogous as for the output label $y_i$:
We fully expand all possible values for $t_i$,
i.e.~calculate the score $p(t_i{=}t|\dots) \cdot p(y_i^{i-1}, t_1^{i-1} | x_1^T)$
for all possible $t$, for all current active hypotheses,
just as we do for $y$.
%just as we also calculate the score % $p(y_i{=}y|\dots)$
%for all possible $y$, for all active hypotheses.
%
However, we have various options for
when, what and how do we apply \textbf{pruning}.
%E.g.~for all possible $t_i$ given the current active hypotheses,
%we could prune with the same beam size as before.
%But maybe this is not optimal.
%Maybe no pruning at all should be done here,
%but only after we went through all combinations of $p_i$ and $t_i$
%for all current active hypotheses.
%
After we calculated the scores for $y_i$,
based on the joint scores $p(y_i{=}y |\dots) \cdot p(y_i^{i-1}, t_1^i | x_1^T)$,
we \textbf{prune} to the best $K_y$ hypotheses.
Now on the choice of $t_i$,
we have $K \cdot T'$ active hypotheses,
where $K=1$ at the beginning, and then $K=K_y$.
For pruning the hypotheses on $t_i$
based on the joint scores $p(t_i{=}t|\dots) \cdot p(\dots)$:
\begin{itemize}
\item We either select the overall $K_t$ best,
ending up with $K_t$ hypotheses.
\item Or for each previous hypotheses (last choice on $y_{i-1}$),
we select the best $K'_t$ choices for $t_i$,
ending up with $K_t = K \cdot K'_t$ hypotheses.
I.e.~we \textbf{only expand} and don't prune $y_1^{i-1}$ hypotheses away at this point,
but only prune away new choices of $t_i$,
i.e.~take the top-$K'_t$ of possible $t_i$ values.
\end{itemize}

% trecombtrain, e.g.: hard-att4j.trecombtrain.alignep180.attwl01.base2.conv2l.specaug4a
As a further approximation,
we can also discard hypotheses in the beam which have the same
history $y_1^i$ but a different history $t_1^i$.
We currently do this for the Viterbi alignment during training with ground truth,
such that we only consider
\[ t_i \mapsto \max_{t_1^{i-1}} p(t_i | \dots) \cdot p(y_1^{i-1}, t_1^{i-1} | x_1^T) . \]
We do not use this for decoding yet.
In general, this can make more efficient use of the beam.

Note that this is in the worst case only slower to the baseline
by some small constant factor.
In practice, in the batched GPU-based beam search which we do,
we do not see any difference.
% base model, GPU 1080: 0:05:23 min for search.o189724.2.gz
% latent model, GPU 1080: 0:05:37 min for search.o187686.2.gz
Once we use different models for $p(t|\dots)$,
we can even potentially gain a huge speedup.

\section{Experiments}

\begin{table}[t]
% Models:
% hard-att-local-win10-imp.ls01.laplace1000.hlr, imports: "filename": "base/data-train/base2.conv2l.specaug4a/net-model/network.160"
%\def\arraystretch{0.9}
\setlength{\tabcolsep}{0.2em}
  \caption{Overall comparisons of models.
  We use the beam sizes $K_y = 12$ and $K_t = 48$
  with \emph{expand} on $t$ as explained in \Cref{sec:decoding}.
  The frame error rate (FER) is calculated on the labels,
  feeding in the ground truth seq., on some cross-validation set.
  The latent models all import the global soft.~att. baseline,
  which screws the effective num.~of epochs though,
  as the model is changed drastically.}
  \label{tab:final-results}
  \label{tab:baselines}
\centerline{
\begin{tabular}{|c|c|c|c|c|c|c|}
\hline
Model & Effect. & \multicolumn{4}{c|}{WER[\%]} & FER \\
& num. & \multicolumn{3}{c|}{Hub5'00} & Hub5'01  & [\%] \\
& ep. & $\Sigma$ & SWB & CH & $\Sigma$ & \\
\hline\hline
% base2.conv2l.specaug4a, ep 160, 15.3%
Global soft. & \phantom{0}33.3 & 15.3 & 10.1 & 20.5 & 14.9 & 8.5 \\
% base2.conv2l.specaug4a.retrain1: ep 163, 14.3%
& \phantom{0}66.7 & \textbf{14.3} & \phantom{0}9.3 & 19.3 & 14.0 & 8.0 \\
% base2.conv2l.specaug4a.retrain2, ep 154, 14.4%
& 100\phantom{.0} & 14.4 & \phantom{0}9.1 & 19.7 & \textbf{13.9} & 8.0 \\
\hline
% hard-segm-prob-imp.ls01.laplace1000, ep 136,15.7%
Segmental & \phantom{0}58.3 & 15.7 & \phantom{0}9.9 & 21.5 & \textbf{14.6} & 7.4 \\
% hard-segm-prob-imp-retrain1.ls01.laplace1000
% todo still running (200 eps), but best after 100:  ep61: 15.7
& \phantom{0}83.3 & 15.7  & 10.1 & 21.3 & 14.9 & 6.9 \\
\hline
%hard-att3.textend.trecombtrain.attwl01.base2.conv2l.specaug4a.ls01.laplace1000, ep 197: 15.8%
Hard att. & \phantom{0}33.3 & 15.8 & 10.3 &21.3&15.4& 7.3 \\
% hard-att-imp3.textend.trecombtrain.attwl01.base2.conv2l.specaug4a.ls01.laplace1000.hlr, ep 145, 14.8%
% hard-att-imp3.textend.trecombtrain.attwl01.base2.conv2l.specaug4a.ls01.laplace1000, ep 100, 15.1%, 7.31%FER
& \phantom{0}50\phantom{.0} & 15.1 & \phantom{0}9.7 & 20.5 & 14.8 & 7.3 \\
% hard-att-imp3-retrain1.textend.trecombtrain.attwl01.base2.conv2l.specaug4a.ls01.laplace1000, ep 80, 14.4%, 7.03%FER
& \phantom{0}91.7 & \textbf{14.4}  & \phantom{0}9.3 & 19.5& \textbf{14.2} & 7.0 \\
\hline
% hard-att-local-win10.wfb.mdrop.ls01.laplace1000, ep178: 15.4%
Local win. & \phantom{0}33.3 & 15.4 &10.2&20.5&14.9& 8.0 \\
% hard-att-local-win10-imp.ls01.laplace1000.hlr, ep 92, 14.7%, 7.75%FER
& \phantom{0}50\phantom{.0} & 14.7 & \phantom{0}9.4 & 20.0 & 14.1 & 7.7 \\
% hard-att-local-win10-imp-retrain1.ls01.laplace1000.hlr, ep 85: 14.4%
& \phantom{0}83.3 & \textbf{14.4} & \phantom{0}9.1 &19.7& \textbf{13.8} & 7.6 \\
\hline
\end{tabular}}
\end{table}

%\TODO we could also add back our synthetic experiments...
\subsection{Dataset \& global soft attention baseline}

All our experiments are performed on Switchboard \cite{godfrey1992switchboard}
with 300h of English telephone speech.
\ifdefined\isaccepted
Our global soft attention baseline is mostly based on
\cite{zeyer2018:asr-attention,zeyer2018:attanalysis,zeyer2019:trafo-vs-lstm-asr}.
\fi
We use SpecAugment \cite{park2019specaugment} for data augmentation.
We have two 2D-convolution layers,
followed by 6 layer bidirectional LSTM encoder of dimension 1024 in each direction,
with two time-max-pooling layers
which downsample the time dimension by a factor of 6 in total.
Our output labels are 1k BPE subword units \cite{sennrich2015bpe}.
All our experiments are without the use of an external language model.
We use a pretraining/training-scheduling scheme which
% base2.conv2l.specaug4a
\begin{itemize}[itemsep=0pt]
\item starts with a small encoder, consisting of 2 layers and 512 dimensions,
and slowly grows both in depth (number of layers) and width (dimension)
up until the final model size,
\item starts with reduced dropout rates, which are slowly increased,
\item starts without label smoothing, and only later enables it,
\item starts with a less strong SpecAugment,
\item starts with a higher batch size.
\end{itemize}
This scheme is performed during the first 10 full epochs of training.
In addition, for the first 1.5 epochs, we do learning rate warmup,
and after that we do the usual learning rate scheduling \cite{bengio2012practical}.
After every 33.3 epochs, we do a reset of the learning rate.
We show the performance in \Cref{tab:baselines}.
We see that longer training time can yield drastic improvements.
We train these experiments using a single Nvidia GTX 1080 Ti,
and one epoch takes about 3h-6h, depending on the model.

\subsection{Latent variable models}

\subsubsection{Pretraining scheme}

%\TODO looks unfair when comparing effective num of epochs, cannot compare? explain.
% Reviewer: Table 2 showed the comparison between “from scratch” and “imported”.  I think it would be fair to compare these two methods with the similar effective epoch numbers. But in this table, If my understanding is correct, the epoch number  of “from scratch” is 33.3, and that of “imported” is around 50 or 90. We can’t get any conclusions based on such unfair comparison.
\begin{table}[t]
  \caption{From scratch vs imported initialization,
  for hard attention and local windowed soft attention.
  WER on Hub5'00.}
  \label{tab:scratch-vs-import}
\centerline{
\begin{tabular}{|c|c|c|c|}
\hline
Model & Imported baseline, & Effective & WER \\
& num.~of ep. & num.~of ep. &[\%]  \\
\hline \hline
%hard-att3.textend.trecombtrain.attwl01.base2.conv2l.specaug4a.ls01.laplace1000
Hard att. & --- & 33.3 & 15.8 \\
%hard-att-imp3.textend.trecombtrain.attwl01.base2.conv2l.specaug4a.ls01.laplace1000, 15.1%
& 33.3 & 58.3 & 15.1 \\
% hard-att-imp3-retrain1.textend.trecombtrain.attwl01.base2.conv2l.specaug4a.ls01.laplace1000
& 66.7 & 91.7 & 14.4 \\
\hline
%hard-att-local-win10.wfb.mdrop.ls01.laplace1000
Local win. & --- & 33.3 & 15.4 \\
%hard-att-local-win10-imp.wfb.mdrop.ls01.laplace1000, ep 100, 15.1%
& 33.3 & 50\phantom{.0} & 15.1 \\
\hline
\end{tabular}}
\end{table}

As all our proposed models are close to the global soft attention baseline,
we can import the model parameters from the baseline and fine tune.
With the aforementioned methods for stable training
such as reusing the previously best alignment,
we are also able to train from scratch, as we show in \Cref{tab:scratch-vs-import}.
Note that the difference in ``from scratch" training to importing a baseline
is somewhat arbitrary here, because the latent variable models also use some
pretraining scheme as outlined in \Cref{sec:training-latent}.
Thus we end up with two different kinds of pretraining schedules,
and ``from scratch" training
uses much less global soft attention during pretraining.
This merely demonstrates that from scratch training is possible,
i.e.~that we can switch to a fully latent model (without global soft attention) already early in training.
The performance difference in these experiments
might also just be due to the different effective number of epochs.
Note that ultimately after the switch from the global soft attention model
to a latent variable model,
we will always see a drop in WER,
as the model is not the same.
The amount of this degradation can be seen in \Cref{tab:drop-after-import}.
In the following experiments we use the best possible scheduling,
which currently imports existing global soft attention models.

\subsubsection{Prune variants in decoding}

\begin{table}
  \caption{Comparison on different decode pruning on $t_i$,
  for a hard attention model.
  We always use the beam size $K_y = 12$.
  \emph{Expand} as explained in \Cref{sec:decoding},
  and $K_t$ is the beam size after the choice on $t_i$.
  In the case of \emph{expand} with $K_t = K_y$,
  we get the deterministic $\argmax_{t_i}$ approach.
%  We also tried $K_t=96$ in all these settings but there is no further improvements.
  WER on Hub 5'00.}
  \label{tab:decode-prune-t}
\centerline{
\begin{tabular}{|c|c|c|c|}
\hline
Model & Expand $t_i$ & $K_t$ & WER[\%] \\
\hline\hline
% hard-att-imp3-retrain1-recog.t{no,}extend{12,24,48,96}.trecombtrain.attwl01.base2.conv2l.specaug4a.ls01.laplace1000, ep 80
%scores/hard-att-imp3-retrain1-recog.textend12.trecombtrain.attwl01.base2.conv2l.specaug4a.ls01.laplace1000.recog.wers.txt
%epoch  80: 14.8
Hard att. & yes & 12 & 14.8 \\
%scores/hard-att-imp3-retrain1-recog.textend24.trecombtrain.attwl01.base2.conv2l.specaug4a.ls01.laplace1000.recog.wers.txt
%epoch  80: 14.5
& & 24 & 14.5 \\
%scores/hard-att-imp3-retrain1-recog.textend48.trecombtrain.attwl01.base2.conv2l.specaug4a.ls01.laplace1000.recog.wers.txt
%epoch  80: 14.4
& & 48 & 14.4 \\
%scores/hard-att-imp3-retrain1-recog.textend72.trecombtrain.attwl01.base2.conv2l.specaug4a.ls01.laplace1000.recog.wers.txt
%epoch  80: 14.4
%scores/hard-att-imp3-retrain1-recog.textend96.trecombtrain.attwl01.base2.conv2l.specaug4a.ls01.laplace1000.recog.wers.txt
%epoch  80: 14.4
& & 96 & 14.4 \\
%scores/hard-att-imp3-retrain1-recog.textend192.trecombtrain.attwl01.base2.conv2l.specaug4a.ls01.laplace1000.recog.wers.txt
%epoch  80: 14.4
\cline{2-4}
%scores/hard-att-imp3-retrain1-recog.tnoextend12.trecombtrain.attwl01.base2.conv2l.specaug4a.ls01.laplace1000.recog.wers.txt
%epoch  80: 14.5
& no & 12 & 14.5 \\
%scores/hard-att-imp3-retrain1-recog.tnoextend24.trecombtrain.attwl01.base2.conv2l.specaug4a.ls01.laplace1000.recog.wers.txt
%epoch  80: 14.4
& & 24 & 14.4 \\
%scores/hard-att-imp3-retrain1-recog.tnoextend48.trecombtrain.attwl01.base2.conv2l.specaug4a.ls01.laplace1000.recog.wers.txt
%epoch  80: 14.4
& & 48 & 14.4 \\
%scores/hard-att-imp3-retrain1-recog.tnoextend72.trecombtrain.attwl01.base2.conv2l.specaug4a.ls01.laplace1000.recog.wers.txt
%epoch  80: 14.4
%scores/hard-att-imp3-retrain1-recog.tnoextend96.trecombtrain.attwl01.base2.conv2l.specaug4a.ls01.laplace1000.recog.wers.txt
%epoch  80: 14.4
& & 96 & 14.4 \\
\hline
% hard-att-local-win10-imp-recog.t{no,}extend{12,24,48,96}.ls01.laplace1000.hlr, ep 92
%scores/hard-att-local-win10-imp-recog.textend12.ls01.laplace1000.hlr.recog.wers.txt
%epoch  92: 14.7
Local win. & yes & 12 & 14.7 \\
%scores/hard-att-local-win10-imp-recog.textend48.ls01.laplace1000.hlr.recog.wers.txt
%epoch  92: 14.7
& & 48 & 14.7 \\
%scores/hard-att-local-win10-imp-recog.textend96.ls01.laplace1000.hlr.recog.wers.txt
%epoch  92: 14.7
& & 96 & 14.7 \\
\cline{2-4}
%scores/hard-att-local-win10-imp-recog.tnoextend12.ls01.laplace1000.hlr.recog.wers.txt
%epoch  92: 14.8
& no & 12 & 14.8 \\
%scores/hard-att-local-win10-imp-recog.tnoextend48.ls01.laplace1000.hlr.recog.wers.txt
%epoch  92: 14.7
&  & 48 & 14.7 \\
%scores/hard-att-local-win10-imp-recog.tnoextend96.ls01.laplace1000.hlr.recog.wers.txt
%epoch  92: 14.7
& & 96 & 14.7 \\
\hline
\end{tabular}}
\end{table}

\begin{table}[t]
  \caption{Comparison on different $\alpha$ for $p(t|\dots)^\alpha$
  and softmax temperature
  for a hard attention model.
  We always use the beam size $K_y=12$, $K_t = 48$,
  and \emph{expand} $t_i$ (\Cref{sec:decoding}).
  WER on Hub 5'00.}
  \label{tab:decode-att-alpha-temp}
\centerline{
\begin{tabular}{|c|c|c|}
\hline
Softmax temp. & $p(t)$ scale $\alpha$ & WER[\%] \\
\hline\hline
% hard-att-imp3-retrain1-recog.textend48.{aw05,aw01,tw05,tw15}.trecombtrain.attwl01.base2.conv2l.specaug4a.ls01.laplace1000
%scores/hard-att-imp3-retrain1-recog.textend48.aw01.trecombtrain.attwl01.base2.conv2l.specaug4a.ls01.laplace1000.recog.wers.txt
%epoch  80: 21.2
1.0 & 0.1  & 21.2 \\
%scores/hard-att-imp3-retrain1-recog.textend48.aw05.trecombtrain.attwl01.base2.conv2l.specaug4a.ls01.laplace1000.recog.wers.txt
%epoch  80: 14.5
& 0.5  & 14.5 \\
% scores/hard-att-imp3-retrain1-recog.textend48.trecombtrain.attwl01.base2.conv2l.specaug4a.ls01.laplace1000.recog.wers.txt 
%epoch  80: 14.4
 & 1.0 & 14.4 \\
%scores/hard-att-imp3-retrain1-recog.textend48.aw15.trecombtrain.attwl01.base2.conv2l.specaug4a.ls01.laplace1000.recog.wers.txt
%epoch  80: 14.6
 & 1.5  & 14.6 \\ \hline
%scores/hard-att-imp3-retrain1-recog.textend48.tw05.trecombtrain.attwl01.base2.conv2l.specaug4a.ls01.laplace1000.recog.wers.txt
%epoch  80: 14.5
0.5 & 1.0 & 14.5 \\ \hline
%scores/hard-att-imp3-retrain1-recog.textend48.tw15.trecombtrain.attwl01.base2.conv2l.specaug4a.ls01.laplace1000.recog.wers.txt
%epoch  80: 14.6
1.5 & 1.0 & 14.6 \\
\hline
\end{tabular}}
\end{table}

We analyse the different kinds of pruning on the latent variable $t_i$ during decoding,
and the beam size $K_t$, as explained in \Cref{sec:decoding}.
Results are collected in \Cref{tab:decode-prune-t}.
The results indicate that we need a large enough beam size for optimal performance.
%but larger than that has not a big influence anymore.
For the hard attention model,
a simple deterministic decision on $t_i$ is slightly worse than
doing beam search on it.
It does not seem to matter too much for the local windowed attention model,
which probably can recover easily if the choice on $t_i$ is slightly off.
Note that this is very much dependent on the quality of the model $p(t|\dots)$.
If that model is weak, it should be safer and better to use the \emph{expand} prune variant
with some $K_t > K_y$,
as it would not prune away possibly good $y_1^i$ hypotheses.
However, our experiments do not demonstrate this yet,
which could be simply due to an already powerful model.
%\TODO explain more, why. table does not show that.
In our experiments $K_y=12$, $K_t = K_y \cdot K'_t = 48$, i.e.~$K'_t=4$ seems to be enough.
That means that we effectively consider always the top 4 scored choices for $t_i$.
% we also did larger beam on y (same for baseline then...). no difference.
We also did experiments on overall larger beam sizes for $K_y$ as well
but do not see any further improvement.
We also experimented with an exponent on the probability, i.e.~$p(t|\dots)^\alpha$,
but $\alpha \approx 1$ seems to perform best.
We also tried different $\operatorname{softmax}$ temperature factors for $p(t|\dots)$, but again $1$ seems to be the optimum.
These experiments are shown in \Cref{tab:decode-att-alpha-temp}.
We note again that this behaviour is likely dependent on the specific kind of model $p(t|\dots)$.

\subsubsection{Overall results}

We collect the final results in \Cref{tab:final-results}.
%\TODO effective num epochs more clear...
%\TODO also mention computation time for decoding (maybe training as well).
We again also report the total number of epochs of the training data
which were seen. % for a more fair comparison,
%However, in the case of latent attention, although this comparison is a bit screwed by the fact
%that we drastically change the model during training
%in the latent model cases.
We also report the frame error rate (FER),
which we found interesting, as it is consistently much better
for all of the latent models.
Our assumption is that the alignment procedure works well
and it makes a good prediction of the output label easier,
given the good position information.
%given a good fixed (maybe hard) position.
However, we do not see this for the WER.
The segmental model seems to have the highest FER,
while it performs not as good as the other latent models.
This could be due to some overfitting effect, but needs further studying.
We assume this is due to the exposure bias in training,
where it always has seen the ground truth label sequence
and a good alignment.
The global soft attention model has less problems with this,
as there is never a fixed alignment.
Overall, the WERs of the latent models are competitive
to global soft attention.

\subsubsection{Restriction on the max step size $t_i - t_{i-1}$}

\begin{table}[t]
  \caption{Comparison on different maximum step sizes for $p(t|\dots)$.
  WER on Hub 5'00.
  In the unlimited case, there can be step sizes over 150 frames (9s),
  although the mean step size is about 4.3 $\pm$ 4.0 frames (0.26s $\pm$ 0.24s).
  }
  \label{tab:max-fwd-step}
\centerline{
\begin{tabular}{|c|c|c|c|}
\hline
Model & Max.~step size & WER[\%] \\
\hline\hline
% hard-att-imp3-retrain1.fwd10.textend.trecombtrain.attwl01.base2.conv2l.specaug4a.ls01.laplace1000
Hard att. & 10 (0.6s) & 18.1 \\
% hard-att-imp3-retrain1.fwd20.textend.trecombtrain.attwl01.base2.conv2l.specaug4a.ls01.laplace1000
& 20 (1.2s) & 14.7 \\
% hard-att-imp3-retrain1.fwd30.textend.trecombtrain.attwl01.base2.conv2l.specaug4a.ls01.laplace1000
& 30 (1.8s) & 14.5 \\
% hard-att-imp3-retrain1.textend.trecombtrain.attwl01.base2.conv2l.specaug4a.ls01.laplace1000
& $\infty$ & 14.4 \\
\hline
% hard-att-local-win10-imp-retrain1.fwd10.ls01.laplace1000.hlr
Local win. & 10 (0.6s) & 14.6 \\
% hard-att-local-win10-imp-retrain1.fwd20.ls01.laplace1000.hlr
& 20 (1.2s) & 14.5 \\
% hard-att-local-win10-imp-retrain1.fwd30.ls01.laplace1000.hlr
& 30 (1.8s) & 14.5 \\
% hard-att-local-win10-imp-retrain1.ls01.laplace1000.hlr
& $\infty$ & 14.5 \\
\hline
\end{tabular}}
\end{table}

In all the experiments, the model $p(t|\dots)$ operates offline on the whole input sequence,
and then calculates a $\operatorname{softmax}$ over the time dimension.
For an online capable model, we cannot use $\operatorname{softmax}$ for the normalisation
unless we restrict the maximum possible step size, which is maybe a good idea in general.
Specifically, $p(t_i{=}t|\dots) = 0$ for all $t > t_{i-1} + \Delta_t$ for some $\Delta_t$.
We collected the results in \Cref{tab:max-fwd-step}.
We can conclude that a maximum step size of 30 frames, which corresponds to 1.8 seconds, is enough,
although this might depend on the dataset.

\subsubsection{Further discussion}

\begin{table}
\setlength{\tabcolsep}{0.1em}
  \caption{Our incremental effort to tune the hard attention model.
  WER on Hub5'00.
  We report the WER directly after importing the baseline (without any update),
  for training begin (50h of training data),
  and then after that the best ($\ge$100h of training data).
  A connected cell in the table means that this number
  cannot have changed conceptually.}
  \label{tab:dev-history}
  \label{tab:drop-after-import}
\centerline{
\begin{tabular}{|l|c|c|c|c|}
\hline
Incremental changes: & \multicolumn{4}{c|}{WER[\%]} \\
All changes from top to bottom & base- & after & train & further \\
add up. &line& imp. & begin & best \\
\hline \hline
%hard-att-imp.base2.conv2l.specaug4a, ep1: 18.7%, ep7: 17.0%, ep 78: 19.1%
% (lr 0.001, lr warmup)
Starting point: no label smoothing, & 15.3 & 18.7 & 17.0 & 19.1 \\
no weight feedback, $K_y = K_t = 12$ &&&& \\
\cline{1-1}\cline{3-5}
%hard-att-imp2.base2.conv2l.specaug4a, ep1: 16.5%, ep7: 16.9%, ep 36: 16.9%
% diff to hard-att-imp: lower lr 0.0001, no lr warmup, using weight feedback
Weight feedback (soft att.~weights), && 16.5 & 16.9 & 16.9 \\
lower learning rate, no lr.~warmup && & & \\
\cline{1-1}\cline{3-5}
%hard-att-imp3.textend.trecombtrain.attwl01.base2.conv2l.specaug4a, ep1: 16.4%, ep7: 16.9% ep30: 16.8%
% diff to hard-att-imp2: t_recomb_train, att weight loss scale 0.1, t_extend, K_t = 48, accum_att_weights uses hard attention
Approximated recombination for && 16.4 & 16.9 & 16.8 \\
realignment, \emph{expand} on $t$, $K_t=48$, &&&& \\
$\log p(t|...)$ loss scale 0.1, &&&& \\
accum.~att.~weights use hard att. &&&& \\
\cline{1-1}\cline{4-5}
%hard-att-imp3.textend.trecombtrain.attwl01.base2.conv2l.specaug4a.ls01, ep1: 16.4%, ep7: 16.4%, ep42: 16.2%
Label smoothing ($0.1$) & &&  16.4 & 16.2 \\
\cline{1-1}\cline{5-5}
%hard-att-imp3.textend.trecombtrain.attwl01.base2.conv2l.specaug4a.ls01.laplace1000, ep1: 16.4%, ep7: 16.4%, ep100: 15.1%
Correct seq.~shuffling & &&  & 15.1 \\
\cline{1-1}\cline{5-5}
%hard-att-imp3.textend.trecombtrain.attwl01.base2.conv2l.specaug4a.ls01.laplace1000.hlr, ep1: 16.4%, ep7: 16.4%, ep145: 14.8%
Higher learning rate, lr.~warmup & & & & 14.8 \\
\hline\hline
%hard-att-imp3-retrain1.textend.trecombtrain.attwl01.base2.conv2l.specaug4a.ls01.laplace1000, ep1: 15.4%, ep7: 15.3%, ep80: 14.4%
Import better baseline & 14.3 & 15.4 & 15.3 & 14.4 \\
\hline
\end{tabular}}
\end{table}

Achieving to the final results required multiple states of tuning.
In \Cref{tab:dev-history}, we share our development history up to the final results.
The starting point does not use label smoothing,
which was actually a bug and not intended.
We see that label smoothing has a big positive effect.
Even bigger was the effect by not shuffling the sequences, which was a bug as well.
As we can see from the table, this makes an absolute WER difference of more than 1\%.
While it was never the intention to do that,
we think it might still be interesting
for the reader to follow the history of our development.
We see that in our initial experiments,
the training actually made the model worse,
which can be explained by running into a bad local optimum
due to an unstable alignment procedure.
Fortunately the alignment procedure becomes more stable for the further experiments.
The global soft attention baseline also used weight feedback,
and it was not obvious whether the latent models should have that as well,
and how exactly.
We ended up in the variant that we use the hard choice on $t_i$
instead of the attention weights,
and create a ``hard" variant of accumulated weights,
although it is not fully clear whether this is the optimal solution.

\begin{figure}[t]
\centering
\includegraphics[width=1.03\linewidth]{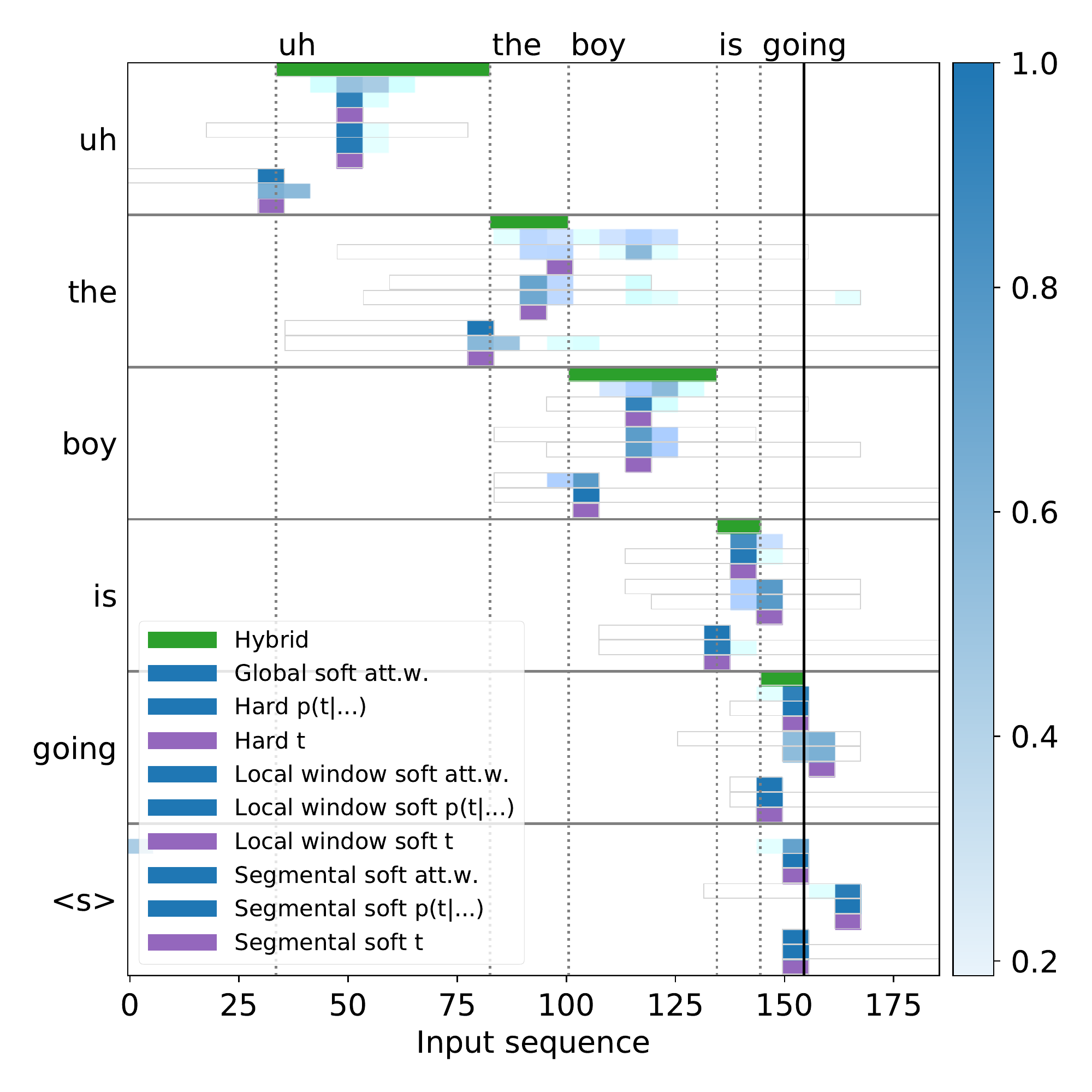}
\vspace{-0.8cm}
\caption{This figure shows the soft and hard attention weights, or segment boundaries
of all the models,
including a reference Viterbi alignment from a hybrid NN/HMM model.
The light grey boxes mark the area where the attention weights or probability distribution
is non-zero.
Some of the models define the attention weights on a longer sequence,
due to the downscaling, or additionally padded zeros at the end.}
\label{fig:att-weights}
\end{figure}

In \Cref{fig:att-weights} we can see the different alignment behaviour
and attention weights of each model.
We can see that most models behave as intended.
In this direct comparison, we see that the global soft attention is noticeably less sharp.

\section{Conclusion \& future work}
We introduced multiple monotonic latent attention models
and demonstrated competitive performance to our strong global soft attention baseline.
From the low FER, we speculate that we have a stronger exposure bias problem now
(not only the ground truth labels $y_1^{N-1}$ but also the time alignment $t_1^N$).
This problem can be solved e.g.~via more regularisation
or different training criteria
such as minimum WER training \cite{prabhavalkar2018minwer}
and we expect to get future improvements.
By further tuning, eventually we expect to get consistently better than the global soft attention baseline.
Future work will also include different alignment models as well as online capable encoders.
Also, while we do label-synchronous decoding in this work,
a latent variable model allows for time-synchronous decoding as well,
which is esp.~interesting the the context of online streaming.

\ifdefined\isaccepted
\newpage
\section*{Acknowledgements}
% http://www-i6.informatik.rwth-aachen.de/i6wiki/Acknowledgments
\begin{wrapfigure}[2]{l}{0.082\textwidth}
   \vspace{-7mm}
    \begin{center}
        \includegraphics[width=0.11\textwidth]{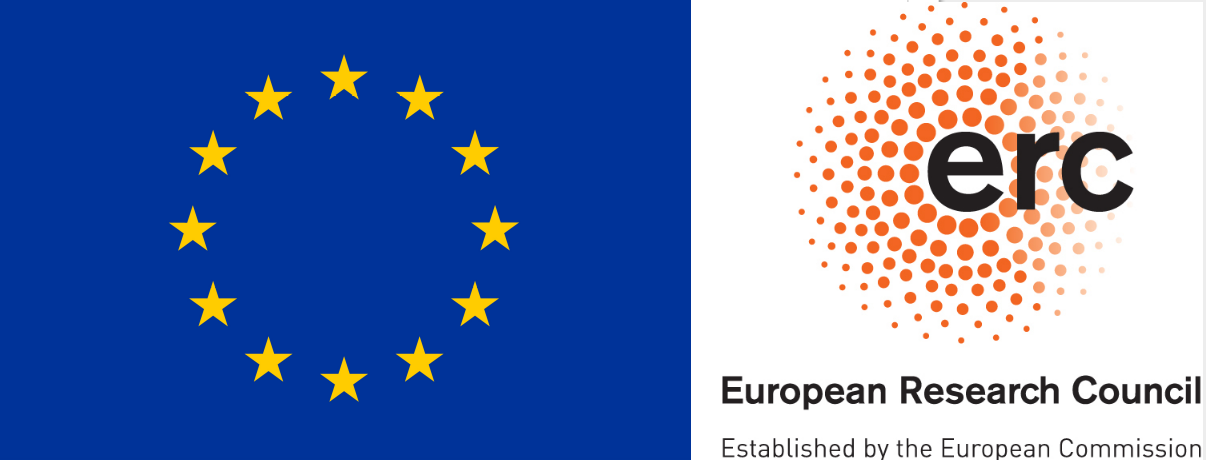} \\
    \end{center}
    \vspace{-12mm}
\end{wrapfigure}
%\footnotesize
%\scriptsize
%\setstretch{0.4}
This work has received funding from the European Research Council (ERC)
under the European Union’s Horizon 2020 research and innovation programme
(grant agreement No 694537, project "SEQCLAS")
and from a Google Focused Award.
The work reflects only the authors’ views
and none of the funding parties is responsible
for any use that may be made of the information it contains.
%\normalsize
\fi

%\vfill
%\pagebreak


%\bibliographystyle{IEEEbib}
\bibliography{refs}
\bibliographystyle{icml2020}

\end{document}